\documentclass[letterpaper, 10 pt, conference]{ieeeconf}  % Comment this line out if you need a4paper

\IEEEoverridecommandlockouts

\usepackage{graphicx} % for pdf, bitmapped graphics files
\usepackage{amsmath} % assumes amsmath package installed
\usepackage{amssymb}  % assumes amsmath package installed
\usepackage[dvipsnames]{xcolor}
\usepackage{xspace}
\usepackage{url}

\usepackage{array}
\usepackage{booktabs}

\usepackage{hyperref}
\usepackage{subfigure}
\usepackage{bytefield}
\usepackage{multirow}
\usepackage{tikz}
\usepackage{tikz-timing}[2014/10/29]
\usetikztiminglibrary[rising arrows]{clockarrows}
\usepackage{rotating}
\usetikzlibrary{positioning, shapes, arrows, automata}
\usepackage{glossaries}
\glsdisablehyper % Do not create hyperlinks for gls
\usepackage[normalem]{ulem}

\usepackage{xparse} % NewDocumentCommand, IfValueTF, IFBooleanTF

\NewDocumentCommand{\busref}{som}{\texttt{%
#3%
\IfValueTF{#2}{[#2]}{}%
\IfBooleanTF{#1}{\#}{}%
}}

\usepackage{marginnote}

\hypersetup{
    colorlinks,
    linkcolor={red!50!black},
    citecolor={blue!50!black},
    urlcolor={blue!80!black}
}

\newif\ifdraftcolor
\ifdraftcolor
\newcommand{\fer}[1]{{\color{orange}#1}}
\newcommand{\longshot}[1]{}
\newcommand{\fixme}[1]{{\color{red}FIXME: #1}}

\newcommand{\deletecamready}[1]{{\color{Red}\sout{#1}}}
\newcommand{\authorlist}{Authors}
\else
\newcommand{\fer}[1]{#1}
\newcommand{\longshot}[1]{}
\newcommand{\fixme}[1]{}

\newcommand{\deletecamready}[1]{}
\newcommand{\authorlist}{Fernando Cladera, Kenneth Chaney,\\
M. Ani Hsieh, Camillo J. Taylor, and Vijay Kumar}
\fi

\newacronym{cots}{COTS}{commercial-off-the-shelf}
\newacronym{uav}{UAV}{unmanned aerial vehicle}
\newacronym{ugv}{UGV}{unmanned ground vehicle}
\newacronym{grd}{GRD}{ground resolution distance}
\newacronym{sfm}{SFM}{structure from motion}
\newacronym{hdr}{HDR}{high dynamic range}
\newacronym{fov}{FoV}{field of view}
\newacronym{agl}{AGL}{above ground level}
\newacronym{erc}{ERC}{event rate controller}
\newacronym{psnr}{PSNR}{peak signal-to-noise ratio}
\newacronym{ssim}{SSIM}{structural similarity index}
\newacronym{odm}{ODM}{OpenDroneMap}
\newacronym{gnss}{GNSS}{Global navigation satellite system}
\newacronym{utm}{UTM}{Universal Transverse Mercator}

\newcommand{\papertitle}{EvMAPPER\xspace}

\title{\LARGE \bf
\papertitle: High Altitude Orthomapping with Event Cameras
}

\author{\authorlist%
\thanks{All authors are with GRASP Laboratory, University of Pennsylvania.
Corresponding author: {\tt\small fclad@seas.upenn.edu}.}
\thanks{
We gratefully acknowledge the support of
ARL DCIST CRA W911NF-17-2-0181, % DCIST (multirobot)
NIFA grant 2022-67021-36856, %anything to do with agriculture/forestry
the IoT4Ag Engineering Research Center funded by the National Science Foundation (NSF) under NSF Cooperative Agreement Number EEC-1941529, %anything to do with agriculture/forestry
and NVIDIA. 
}
}

\begin{document}

\maketitle
\thispagestyle{empty}
\pagestyle{empty}

%%%%%%%%%%%%%%%%%%%%%%%%%%%%%%%%%%%%%%%%%%%%%%%%%%%%%%%%%%%%%%%%%%%%%%%%%%%%%%%%
\begin{abstract}
Traditionally, \glspl{uav} rely on CMOS-based cameras to collect images about the world below. 
One of the most successful applications of UAVs is to generate orthomosaics or orthomaps, in which a series of images are integrated together to develop a larger map. 
However, the use of CMOS-based cameras with global or rolling shutters mean that orthomaps are vulnerable to challenging light conditions, motion blur, and high-speed motion of independently moving objects under the camera. 
Event cameras are less sensitive to these issues, as their pixels are able to trigger asynchronously on brightness changes. 
This work introduces the first orthomosaic approach using event cameras.
In contrast to existing methods relying only on CMOS cameras, our approach enables map generation even in challenging light conditions, including direct sunlight and after sunset.
\fixme{We demonstrate the capability of event cameras to generate orthomosaics even under fast flight conditions, reaching speeds of up to 9 m/s.}
\fixme{Add metrics of performance.}
\longshot{Moreover, we can segment independently moving objects from the background scene, enabling tracking of moving objects in the orthomap.}
\end{abstract}

The source code for \papertitle, the high-altitude hardware, and the dataset collected in this paper are available open source\footnote{\url{https://evmapper.fcladera.com}}.

%%%%%%%%%%%%%%%%%%%%%%%%%%%%%%%%%%%%%%%%%%%%%%%%%%%%%%%%%%%%%%%%%%%%%%%%%%%%%%%%

\glsresetall

\section{Introduction}
High-altitude photography has proven useful for surveillance, mapping and inspection applications. High-vantage points can provide information that is not easily accessible by ground imaging. Additionally, high-altitude photography offers superior \gls{grd} compared to satellite imagery.
This advantage is due to lower flight altitude, the limitations imposed by the Rayleigh criterion limit~\cite{valenzuela2019basic}, and reduced effects of atmospheric conditions~\cite{jacobsen2005high}. 

High-altitude images are usually integrated together into orthomosaics by
projecting the images into a planar map. Orthomosaics are generated with proven
computer vision techniques, such as \gls{sfm} and mesh reconstructions. There
are multiple \gls{cots} software packages for creating orthomosaics, such as
Pix4D~\cite{pix4dProfessionalPhotogrammetry} and
\gls{odm}~\cite{opendronemapDroneMapping}. The orthomosaic's quality directly depends on the quality of the input images. Generating orthomosaics is an offline
process that is prone to failures if the data is not properly conditioned. For
example, blurry images may lead to fewer features and hinder the feature matching
process. Similarly, low overlap between images, strong winds, or vibrations can
also lead to failures~\cite{opendronemapTutorialsx2014}.

\begin{figure}
    \centering
    \includegraphics[width=\linewidth]{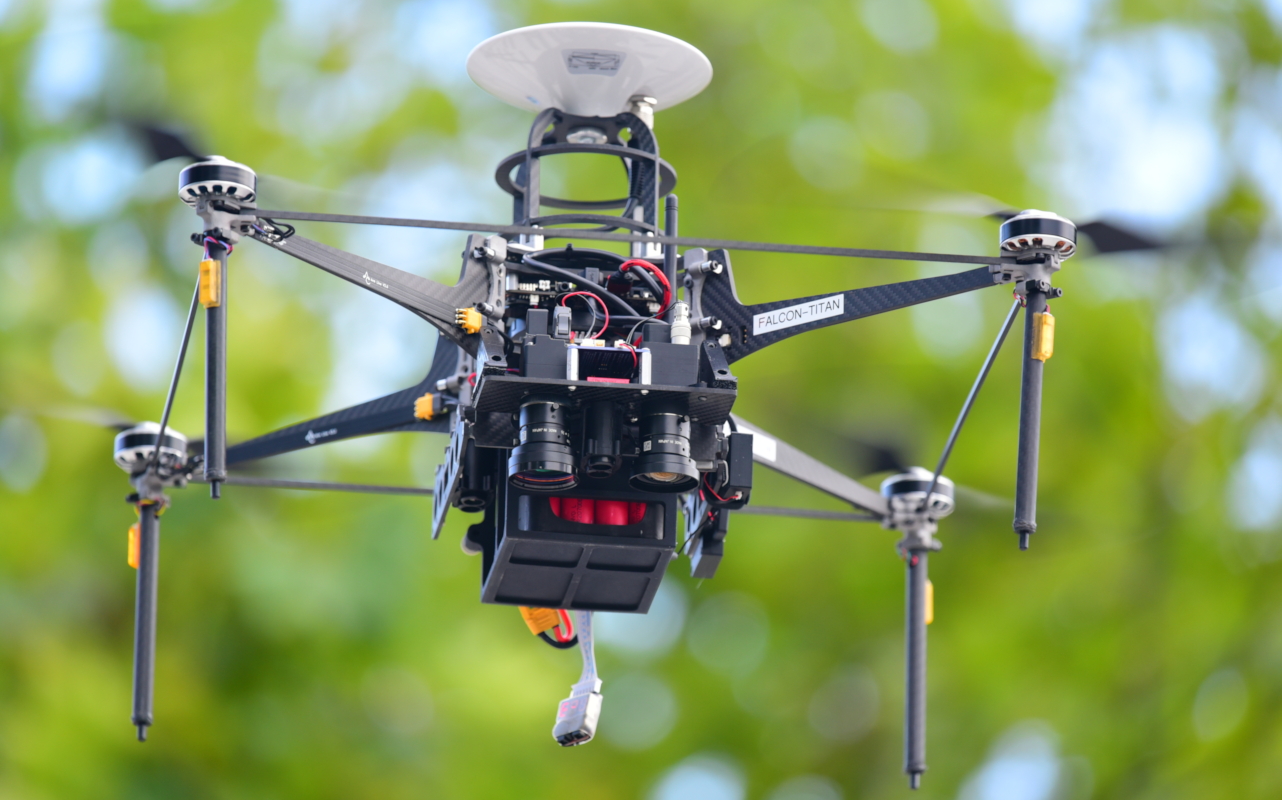}

    \vspace{.1cm}
    \includegraphics[width=.49\linewidth]{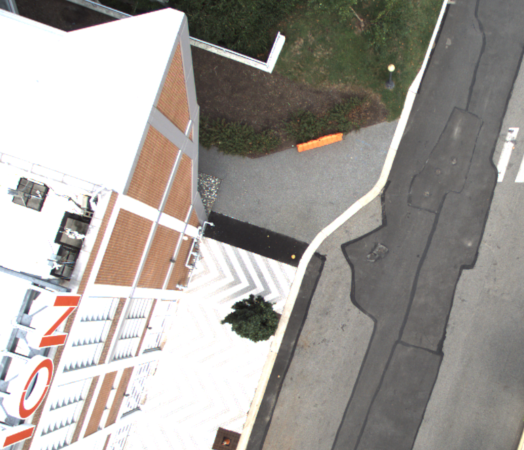}
    \includegraphics[width=.49\linewidth]{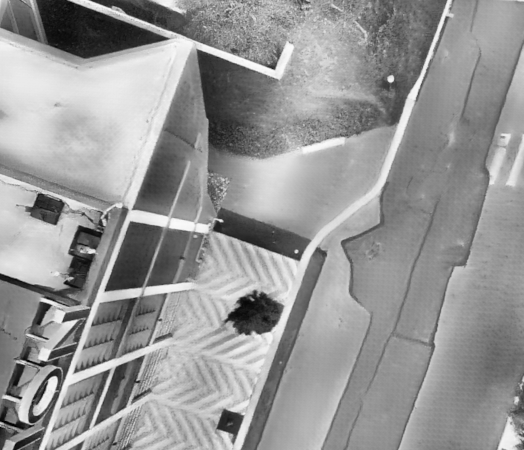}

    \caption{\uline{Top}: The Falcon 4 aerial platform used for high-altitude experiments. The sensor stack, equipped with an IMU, a range sensor, an RGB camera, and an event camera, was mounted at the front. \uline{Bottom left}: artifacts of CMOS-based cameras in high-altitude photography: the sidewalk is washed out due to high brightness. \uline{Bottom right}: reconstructed frame with event cameras displaying higher level of detail in challenging light conditions.
    }
    \label{fig:fig1}
    \vspace{-.5cm}
\end{figure}

Some of the issues encountered during orthomosaic creation stem from the limitations of the sensors employed. CMOS-based cameras are susceptible to motion blur due to wind or \gls{uav} vibration, particularly when long exposures are used. Their limited dynamic range makes it difficult to capture regions with both high brightness and shadows. Fast movements under the camera, resulting from high flight speeds, can lead to failed feature matches for \gls{sfm}. Finally, rolling-shutter CMOS-based cameras are susceptible to  distortion due to the linear readout of pixels. Some of the examples of these issues are observed in Fig.~\ref{fig:fig1}.
There are strengths to CMOS-based cameras and to overcome the above limitations we seek methods to replace and/or augment existing CMOS-based camera data. 

Event cameras are increasingly used in robotics to overcome the limitations of CMOS-based cameras, thanks to their superior dynamic range, higher temporal resolution, and reduced motion blur. Recently, event cameras have seen a significant increase in resolution, nearing the 1 Megapixel mark.
This resolution increase makes event cameras useful for applications where high detail is required. While some works have used event cameras for high-altitude HDR~\cite{li2024ers} imaging or low-light navigation~\cite{escudero2023enabling}, no prior work has explored the use of event cameras for high-resolution orthomosaics. We believe that event cameras can be leveraged to overcome the limitations of traditional imaging sensors in various robotics applications, particularly given the recent success of high-altitude imaging using \glspl{uav}.
However, to employ event cameras for use in orthomosaic mapping, we require a sensor stack which can collect time synchronized data to allow event fusion. Additionally, we need processing methods to convert the collected event data into a format compatible with existing open source orthomosaic mapping tools. 

\textbf{Our contributions are}:
\begin{itemize}
    \item The development of a hardware and software architecture to capture synchronized high-resolution events, RGB images, IMU measurements, and range measurements. 
    \item \fer{A method to integrate the event camera data into off-the-shelf orthomosaic generation tools, and benchmark the results against RGB reconstructions.}   
    \item An open-source high-altitude event camera dataset, comprising synchronized data of our high-altitude \gls{uav} flying in challenging light conditions and at high speed. 
    \end{itemize}
Our contributions demonstrate that event-camera orthomapping is a promising direction of research, and our dataset provides vital baselines for future researchers in the field.  
\section{Related Work}
The most relevant research related to our work can either be classified as using event cameras to increase the dynamic range of RBG cameras or leveraging them for event reconstruction. We briefly summarized this existing work.
%\subsection{High Dynamic Range with RGB and Events}

Event cameras feature a high dynamic range as each pixel in the sensor can trigger independently of the other ones. Some works have leveraged event cameras to increase the dynamic range of CMOS-based cameras. In \cite{scheerlinck2018continuous}, the authors pioneered the use of mixed representations, fusing image frames and events. 
In~\cite{haoyu2020learning}, the authors used event cameras to deblur and enhance a video stream before upsampling. 
In~\cite{messikommer2022multi}, the authors proposed a network architecture using deformable convolutions and LSTM to enhance the dynamic range in static images. The authors used a similar set of sensors to the ones used in this work mounted on a beam-splitter setup. More recently, Li et al.~\cite{li2024ers} recognized that event cameras can be used to enhance the dynamic range of aerial images. The authors propose a gradient-enhanced \gls{hdr} reconstruction network coupled with an event-based dynamic range enhancement network.
While these existing works highlight the advantages of event cameras for aerial imaging, they do not address the integration of the sensor for mapping.

%\subsection{Event Reconstruction}
Multiple works have focused on reconstruction of brightness frames from event data. Reconstructed images could be used to apply mature computer vision algorithms~\cite{gallego2020event}. We apply this approach in our work, leveraging existing tools for orthomapping. Besides higher dynamic range, reconstructed frames also have a higher temporal resolution, allowing for high-speed video reconstruction~\cite{rebecq2019high}. Event reconstruction can be performed using filter-based methods~\cite{scheerlinck2018continuous}, pixel-wise integration~\cite{brandli2014real, pfrommer2024recon, bisulco2021fast}, or learning methods. E2VID~\cite{rebecq2019high, Rebecq19cvpr} proposed a UNet-like architecture to synthesize networks from events, with outstanding results. As we are interested in textures of the environment, we rely on event reconstruction as part of our pipeline.
\section{Methods}
\label{sec:methods}
To address the problem of using event cameras for orthomosaic mapping, we need time-synchronized sensor data. We also need to preprocess this data to adapt it to existing orthomosic mapping frameworks.  

\begin{table*}[t]
    \centering
    \begin{tabular}{p{1cm}p{1.5cm}p{0.5cm}p{2cm}p{1.5cm}p{1.5cm}p{1.5cm}p{2.2cm}p{1.5cm}}
        Sequence  & Duration [s] & Area & Time of the day & Height [m] & Speed [m/s]& Bias on/off& Overlap [\%]& Illumination\\
       \hline
       \texttt{F1.D.1}  & 514   & A &  Noon       & 40 & 3  &  0/0     &  82\%  & Cloudy   \\
       \texttt{F1.D.2}  & 507   & A &  Noon       & 40 & 3  &  50/50   &  82\%  & Cloudy   \\
       \texttt{F2.D.1}  & 615   & A &  Afternoon  & 40 & 3  &  50/50   &  64\% & Sunny    \\
       \texttt{F2.D.2}  & 614   & B &  Afternoon  & 40 & 3  &  100/100 &  64\% & Sunny    \\
       \texttt{F2.D.3}  & 528   & A &  Evening    & 40 & 3  &  50/50   &  64\% & Sunny    \\
       \texttt{F2.D.4}  & 541   & A &  Evening    & 40 & 3  &  0/0     &  64\% & Sunny    \\
       \texttt{F2.N.1}  & 555   & A &  Sunset     & 40 & 3  &  0/0     &  64\% & -        \\
       \texttt{F2.N.2}  & 554   & A &  Dusk       & 40 & 3  &  50/50   &  64\% & -        \\
       \texttt{F2.N.3}  & 541   & A &  Night      & 40 & 3  &  100/100 &  64\% & -        \\
       \texttt{F3.D.1}  & 1282  & A &  Afternoon  & 35 & 3  &  0/0     &  80\% (cross-hatch) & Cloudy   \\
       \texttt{F3.D.2}  & 671   & A &  Afternoon  & 40 & 3  &  0/0     &  82\% & Cloudy   \\
        \texttt{F3.D.3} & 558   & A &  Evening    & 35 & 6  &  0/0     &  80\% & Cloudy   \\
       \texttt{F3.D.4}  & 489   & A &  Evening    & 35 & 9  &  0/0     &  80\% & Cloudy   \\
       \texttt{F3.N.1}  & 832   & A &  Sunset     & 35 & 3  &  0/0     &  80\% & -        \\
       \texttt{F3.N.1}  & 853   & A &  Dusk       & 35 & 3  &  0/0     &  80\% & -        \\
       \hline
    \end{tabular}
    \caption{Data sequences for the high-altitude event camera dataset, showcasing different light conditions, flight patterns, and bias.}
    \label{tab:dataset}
    \vspace{-.5cm}
\end{table*}

\subsection{UAV and Data Acquisition Hardware}
% \fixme{Maybe change this to a table?}
\begin{figure}
    \centering
    \includegraphics[width=\linewidth]{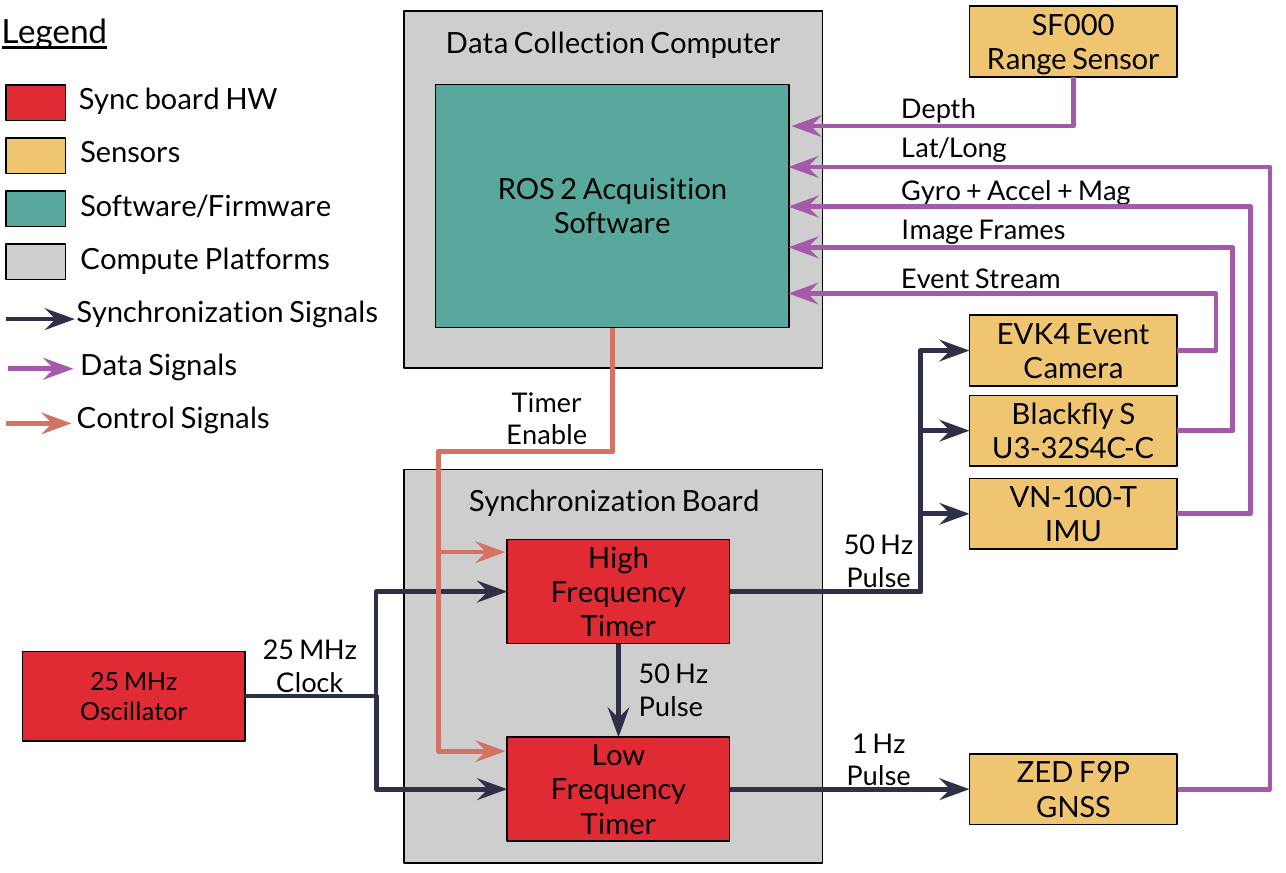}
    \caption{System architecture for data acquisition. The synchronization board generates pulse signals that are used to trigger (Blackfly S) or timestamp sensors (event camera, IMU, GNSS). Data is collected on an onboard computer running ROS 2, and synchronization is performed after the fact. The single-point LiDAR distance sensor is the only sensor that is not hardware synchronized.
    }
    \label{fig:sysdiagram}
    \vspace{-.5cm}
\end{figure}

\subsubsection{Time Synchronization}
Time synchronization is essential for multi-sensor datasets including event cameras, due to the high resolution of the sensors~\cite{Chaney_2023_CVPR, osadcuks2020clock}. In this work, time synchronization was achieved using physical synchronization signals between the different sensors, as shown in Fig.~\ref{fig:sysdiagram}.

A synchronization board based on the STM32L011 microcontroller generates signals using two synchronized timers at $50\,\text{Hz}$ and $1\,\text{Hz}$. The $50\,\text{Hz}$ signal is used to trigger frame captures on the RGB camera. Additionally, the event camera and the IMU use this signal to timestamp their measurements. 
Specifically, on the event camera, every time a signal is received, the local sensor timestamp is recorded and transmitted to the computer as part of the event stream.  
In addition, each IMU message includes the time elapsed between the current measurement and the last synchronization signal. 
A similar timestamping mechanism is used for the \gls{gnss}, recording the \gls{gnss} timestamp when a synchronization signal is received.

\begin{figure}[t]
\centering
%\vspace{.5cm}
\input{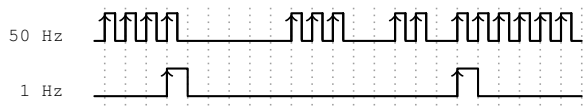}
\caption{Time synchronization pattern inserted in the synchronization signal. The \emph{silence gaps} in the signal are predefined, enabling to identify the beginning of the timing sequence for all the sensors.}
\label{fig:timepattern}
\vspace{-.5cm}
\end{figure}

An important problem we address is matching the first synchronization pulse across sensors. 
To generate an unequivocal \emph{start} signal for all the sensors, we created a temporal pattern in the synchronization signal upon receipt of the \emph{timer\_enable} command by the data collection computer. Data synchronization can be performed after the fact by analyzing the different timestamps of the messages. An example of the temporal pattern can be observed in Fig.~\ref{fig:timepattern}.

\label{subsubsec:time_sync}

\subsubsection{Hardware description}
\label{subsec:hwdescription}
Collecting data for orthomosaic mapping requires building a hardware sensor suite that satisfies the payload capabilities of a UAV, including an event camera and RGB camera.  
As such, we perform experiments with the Falcon 4 platform~\cite{liu2022large}, fitted with a data collection computer with an AMD Ryzen 9 7940HS processor, 32 GB of RAM, and 1 TB SSD. The flight controller is an ARKV6X, capable of flying pre-defined GPS waypoint missions. 

The UAV is fitted with the following sensors:
\begin{itemize}
\item \textbf{FLIR Blackfly S BFS-U3-32S4C-C}: RGB camera, with an IMX252 sensor (2048x1536). The camera is fitted with a Kowa LM5JCM, and captures images at $50\,\text{Hz}$. The \gls{fov} of the camera is $71^\circ\times56^\circ$.
The exposure time for this sensor is adjusted automatically between $5\,\text{ms}$ and $15\,\text{ms}$, using the exposure controller from \texttt{flir\_camera\_driver}~\cite{flirCamDriver}.
\item \textbf{SilkyEVCamHD}: IMX636-based event camera, fitted with a LM5JCM lens. 
We use event camera biases to configure the sensitivity of the sensor to different light conditions. For our experiments, we set all the biases to zero, except \texttt{bias\_diff\_on} and \texttt{bias\_diff\_off} which were set to $0$, $50$ and $100$ for different sequences. The \gls{fov} of the camera is $64^{\circ} \times 39^{\circ}$. In our previous work~\cite{Chaney_2023_CVPR} we observed that using an \gls{erc} negatively impacts the quality of the events. Therefore, we conducted our data collection without employing an \gls{erc}.
\item \textbf{VN-100-T IMU}: 9-DOF temperature compensated IMU, running at $400\,\text{Hz}$. The IMU provides an attitude estimation, compensated acceleration, compensated angular rates, and pressure measurements.
\item \textbf{SF000/B range sensor}: mounted between the two cameras, it provides a single-point depth estimate of the image under the cameras. This sensor captures data at $\approx 60\,\text{Hz}$. The range sensor is the only sensor in our stack that lacks hardware synchronization capability.
This is not a major limitation in this work, as we only use the sensor to determine an adequate altitude to start and stop processing data, as described in Sec.~\ref{subsec:datapreprocessing}.
\item \textbf{ZED-F9P GNSS}: runs at $5\,\text{Hz}$ in multi-constellation configuration. A TOP106 multi-band L1/L2 antenna is used to maximize satellite count and minimize interference from the UAV onboard computer.
\end{itemize}
All these sensors with exception of the GNSS module are fitted in a rigid carbon fiber plate, and attached to the \gls{uav} using vibration-absorbing foam pads. Dampening is critical to reduce high-frequency vibrations from the motors of the UAV, which may generate a significant number of events.
Fig.~\ref{fig:fig1} shows a picture of the platform with the sensor stack, and Fig.~\ref{fig:sysdiagram} presents an overview of the system architecture.

\begin{figure}[b]
    \centering
    \includegraphics[width=\linewidth]{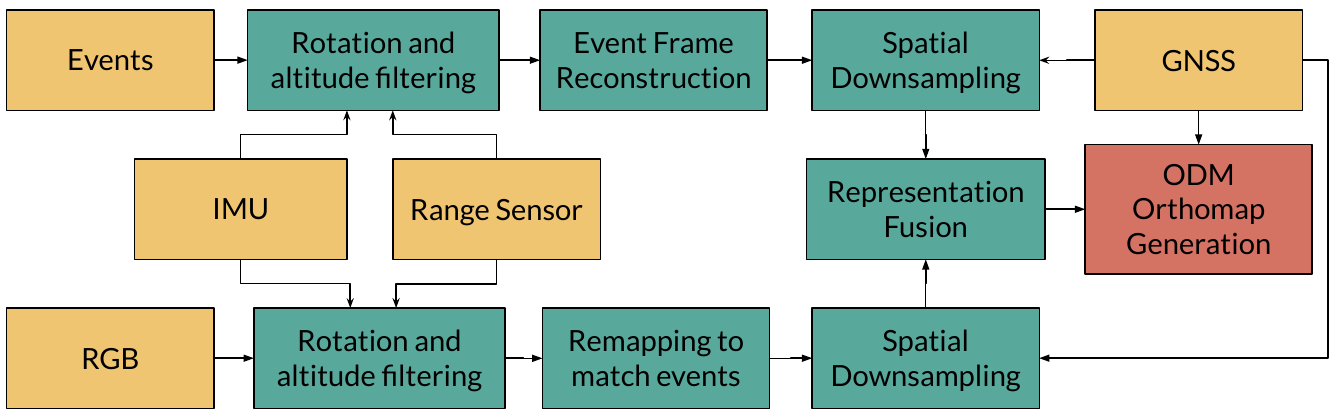}
    \caption{\papertitle data preprocessing pipeline. Events and RGB images are collected when the \gls{uav} is not performing aggressive rotations. Events are reconstructed into frames, and RGB images are remapped to match the event frames.
    The resulting representations are fed into an off-the-shelf orthomap 
    generation tool.}
    \label{fig:pipeline}
\end{figure}

\begin{figure*}[t]
    \centering
    \includegraphics[height=.35\linewidth]{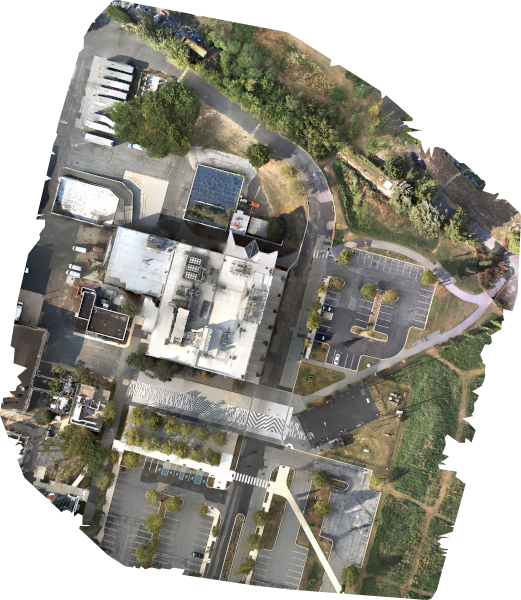}
    \includegraphics[height=.35\linewidth]{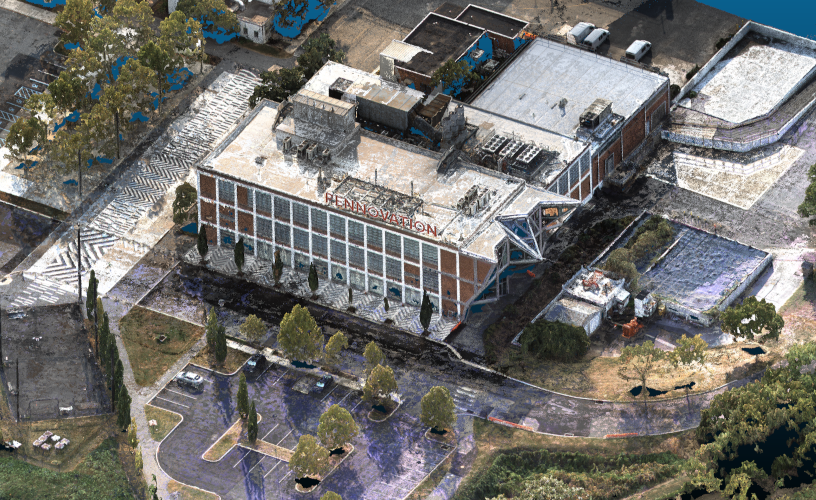}
    \caption{Ground truth reconstruction using \texttt{F3.D.1} and
    \texttt{F3.D.2} sequences. \uline{Left}: Orthophoto. \uline{Right}: Point
    cloud.}
    \label{fig:gt}
\end{figure*}

\subsection{Data Collection Procedure}
\label{subsec:datacollection}
We collected 15 sequences with our high altitude \gls{uav} at the Pennovation campus in Philadelphia, PA, USA. This location showcases a semi-urban environment, with buildings and green areas with grass and trees. The flight area is approximately $11700\,\text{m}^2$. Flights were performed at an altitude of $35\,\text{m}$ or $40\,\text{m}$ \gls{agl}, using pre-defined lawnmower pattern missions.
The details about the sequences are in Tab.~\ref{tab:dataset}. 
The speed of the \gls{uav} varies from $3\,\text{m/s}$ to $9\,\text{m/s}$.
Areas $A$ and $B$ correspond to to different regions of the campus. The bias setting refers to the event camera configuration described in Sec.~\ref{subsec:hwdescription}. Overlap represents the approximate \emph{image overlap} between two parallel flight tracks in the waypoint mission. Finally, the illumination conditions are varied to have enough diversity in the event camera stream.

Sequences are synchronized following the approach described in Sec.~\ref{subsubsec:time_sync} and the resulting synchronized files are saved both in MCAP~\cite{hurliman2024mcap} and \fer{HDF~\cite{folk2011overview}} formats.
Camera and IMU calibrations are obtained using Kalibr~\cite{rehder2016extending}. To calibrate the event and RGB cameras, events are reconstructed using \texttt{simple\_image\_recon}~\cite{pfrommer2024recon}.

\subsection{Data Preprocessing \& Orthomosaic Mapping}
\label{subsec:datapreprocessing}

To ensure the data collected fits within existing orthomosaicing frameworks, we require data preprocessing methods (Fig.~\ref{fig:pipeline}). This includes filtering, reconstructing the events into frame representations, and downsampling and fusion of the RGB and event representations. 
Once data is properly prepared, orthomosaic mapping is performed. 

%\subsubsection{Rotation and Altitude Filtering}
High altitude UAVs are affected by vibrations due to their motors and also wind
gusts. These vibrations generate motion blur in the RGB images, as well as an increased number of events. 
In the context of high-altitude imaging, rotational
vibrations have the largest impact in the quality of the image. Therefore, 
the first stage in \papertitle is to perform rotation filtering of the events
and images using the IMU.
This filtering skips data when the \gls{uav} is performing an aggressive rotation, {\it e.g.}, when the \gls{uav} is accelerating or stopping at a waypoint. 
A threshold is used to eliminate values when the norm of the angular velocity is above $0.4\,\text{rad/s}$. 
Similarly, we want to avoid using data from low-altitude settings when the UAV is taking off or landing. 
We filter out data when the UAV is below $20\,\text{m}$ \gls{agl}.

%\subsubsection{Event Frame Reconstruction}
To reconstruct frames from events we leverage E2VID~\cite{rebecq2019high}.
Reconstructions are performed using $5\,\text{ms}$ event windows. The resulting
images are monochromatic, with higher dynamic range than the RGB counterparts,
yet lacking in low-frequency texture information.

%\subsubsection{Spatial Downsampling}
Orthomosaic reconstructions require adequate spacing between frames: too much
overlap increases the computational cost of the reconstruction, whereas too
little can affect the matching process.
The lateral overlap of the images is defined by the mission flight pattern. The longitudinal overlapping is defined by sub-sampling images and events. 
We use the
\gls{gnss} for this task, to generate a single image every $2\,\text{m}$.

%\subsubsection{Representation Fusion}
The representation fusion uses synchronized RGB frames and the event representation
to generate a new image. This approach is similar to pansharpening, in which a pair of 
multispectral and panchromatic images are fused~\cite{vivione2015critical}.
We selected the Brovey, ESRI, and mean pansharpening approaches to generate the
fused representations and evaluate their results in the generated orthomap.
RGB images are remapped to the event frame reconstruction before processing, using the calibrations obtained in Sec.~\ref{subsec:datacollection}.

%\subsection{Orthomap Generation}
Finally, the input for the orthomap generation is a set of images and \gls{utm} coordinates which can be obtained from \gls{gnss}.
The orthomap generation approaches uses the off-the-shelf reconstruction software
\gls{odm}~\cite{opendronemapDroneMapping}. A resolution of
$1\,\text{cm}/\text{px}$ was chosen for the orthomosaic generation, with an
octree depth of $13$, and a minimum number of features of $12000$.

\begin{figure*}
    \centering
    \includegraphics[width=\linewidth]{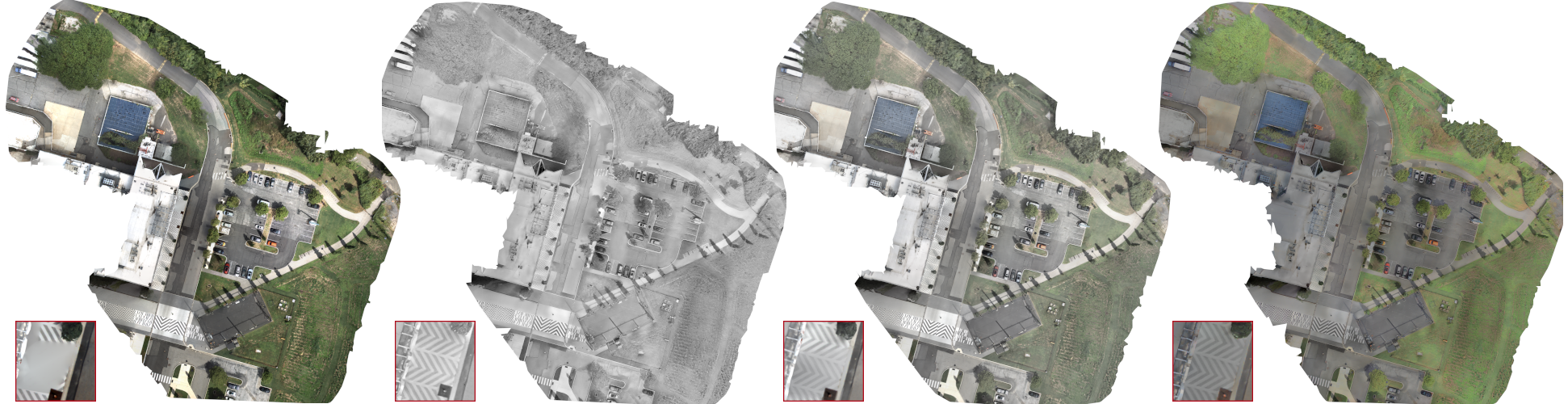}
    \caption{Resulting reconstructions for the \texttt{F1.D.1} sequence for cropped RGB, events only, mean fusion and Brovey fusion. The detail (bottom left rectangle) shows the sidewalk reconstruction, an area of high dynamic range. We observe how the RGB reconstruction fails for this particular sequence, whereas the events and fusions are able to display this area correctly.}
    \label{fig:f1d1}
\end{figure*}
\section{Results}
\label{sec:results}

\begin{table*}[t]
    \centering
    \begin{tabular}{p{1cm}p{3cm}p{2.0cm}p{2.0cm}p{1.5cm}p{2.5cm}p{2.8cm}}
    % {c | c | c | c | c | c}
        Sequence & Type & PSNR Color [dB] & PSNR Gray [dB] & SSIM [\%] & Non-zero pixels [M] \\
        \hline
        \texttt{F1.D.1} & RGB Cropped & \textbf{11.01} & \textbf{10.88} & \textbf{0.54} & 826.58 \\
        \texttt{F1.D.1} & Events Only & 9.41 & 9.39 & 0.49  & 953.39 \\
        \texttt{F1.D.1} & Mean Fusion & 9.87 & 9.77 & 0.50 & 959.39 \\
        \texttt{F1.D.1} & ESRI Fusion & 9.38 & 9.26 & 0.48 & 953.54 \\
        \texttt{F1.D.1} & Brovey Fusion & 9.99 & 9.87 &  0.51 & \textbf{953.73}\\
        %\texttt{F3.D.3} & RGB Cropped  & & & &   \\
        %\texttt{F3.D.4} &  & & & &               \\
        \hline
        \texttt{F3.N.1} &  RGB Cropped & \textbf{12.47} & \textbf{12.41} & 0.63 & 692.33 \\
        \texttt{F3.N.1} &  Events Only & 10.52 & 10.68 & 0.59 & 645.12 \\
        \texttt{F3.N.1} &  Mean Fusion & 12.08 & 12.13 & \textbf{0.65} & 692.09 \\
        \texttt{F3.N.1} &  ESRI Fusion & 9.64  & 9.69  & 0.53 & \textbf{844.95} \\
        \texttt{F3.N.1} &  Brovey Fusion & 10.55 & 10.46 & 0.55 & 844.53 \\
        \hline
        \texttt{F3.N.2} & RGB Cropped & 8.05 & 7.92 & 0.36 &   723.62   \\
        \texttt{F3.N.2} & Mean Fusion & \textbf{9.26} & \textbf{9.12} & \textbf{0.48} &   \textbf{1022.84}  \\
       \hline
    \end{tabular}
    \caption{Results on data sequences for high dynamic range (\texttt{F1.D.1}),
    and low-light conditions (\texttt{F3.N.1} and \texttt{F3.N.2}).\fixme{Results on data sequences for high dynamic range (\texttt{F1.D.1}), high speed conditions (\texttt{F3.D.3} and \texttt{F3.D.3}) and low-light conditions (\texttt{F3.N.1} and \texttt{F3.N.2}). }}
    \label{tab:results}
    \vspace{-.5cm}
\end{table*}

\subsection{Evaluation Approach}

We generate a ground truth reconstruction using a combination of RGB
images from \texttt{F3.D.1} and \texttt{F3.D.2} sequences, without performing any modifications to the images.
This reconstruction corresponds to the highest quality one available in the dataset: both missions are flown in benign light conditions (cloudy evenings), encompassing a larger area than other flights. In addition, \texttt{F3.D.1} flight mission is longer because the \gls{uav} flew a cross-hatch pattern, reducing the probability of non-matched images. The ground truth ortophoto and point cloud can be observed in Fig.~\ref{fig:gt}. 

We align the test image with the ground truth reconstruction using manual feature matching
and homography transformations, using five reference points.
We then compute the \gls{psnr} and
\gls{ssim} metrics between the two images. As the reconstruction may succeed
without including the whole set of images, we also report the total number of
non-zero pixels.
We consider that a reconstruction failed when the reference points for the
homography cannot be found, or when \gls{odm} fails to generate the orthomosaic.

\begin{figure*}
    \centering
    \includegraphics[width=0.37\linewidth]{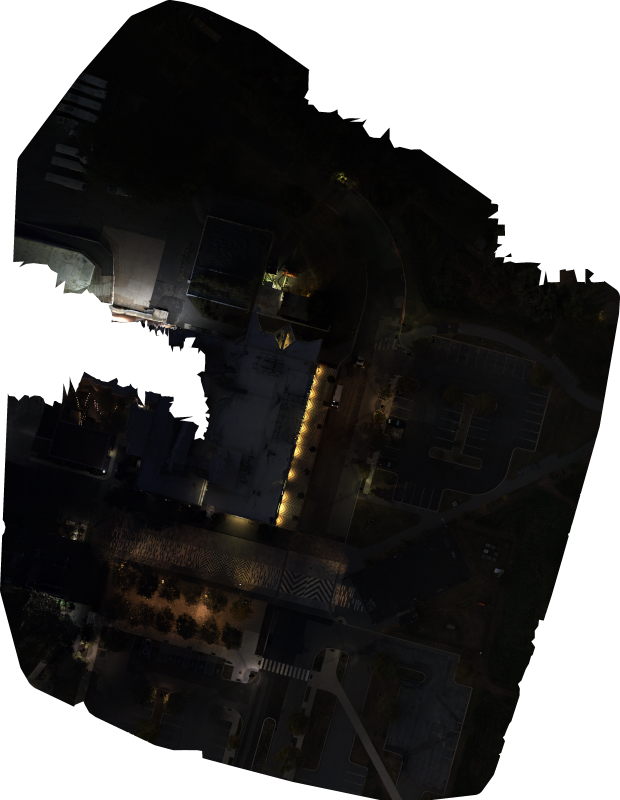}
    \includegraphics[width=0.37\linewidth]{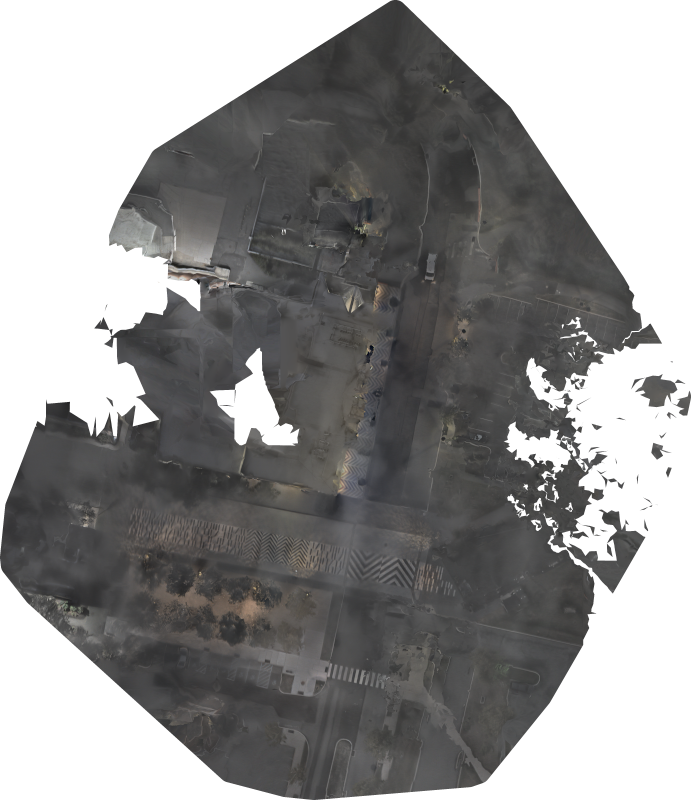}
    \caption{
    Orthomap reconstruction results for the \texttt{F3.N.2} sequence. \uline{Left}: RGB reconstruction. \uline{Right}: mean fusion.
    We observe that the RGB reconstruction has multiple dark areas, while the
    mean fusion reconstruction increases the overall brightness. Some matching
    artifacts can be observed in the event reconstruction.}
    \label{fig:qualitative}
\end{figure*}
\subsection{Results and Discussion}

The results are in Table~\ref{tab:results}.
\emph{RGB cropped} corresponds to the RGB image remapped to the same size as the
event image. \emph{Events only} is the reconstruction of the events without
fusion with RGB. 
For reference, a completely black image would have a PSNR of 7.18
dB, whereas a completely white image would have a PSNR of 6.05 dB.

\subsubsection{High-brightness conditions}
The sequence \texttt{F1.D.1} is a high-brightness sequence as it was recorded
at noon on a sunny day. We can observe in this sequence that the RGB
reconstruction matches the ground truth quite well, achieving the highest \gls{psnr} and \gls{ssim}.
There are, however, some
areas of the image that are better reconstructed using events,  such as the
detail displayed in Fig.~\ref{fig:f1d1}. 
Fused and event representations are able to capture the detail in the sidewalk next to the building, as well as in areas in the shadows.

Among our reconstructions, we observe that mean and Brovey fusion outperform the vanilla event reconstruction. This shows how fusing events with RGB can provide a better reconstruction than using events alone. 

\subsubsection{Low-light conditions}
We can observe that the proposed method performs well on low-light conditions, as shown
by the results in \texttt{F3.N.1} and particularly \texttt{F3.N.2}. For
\texttt{F3.N.1}, the RGB reconstruction still performs slightly better for \gls{psnr}, but the \gls{ssim} is better for the mean fusion. As this sequence is recorded at sunset, there is still 
enough light for the RGB sensor to capture the scene.
On the other hand, the fusion methods performed better for \texttt{F3.N.2}.
It is worth noting that
in this sequence the mean fusion was the only method that was able to
generate results besides the RGB cropped image. The other methods 
(ESRI, events only, and Brovey) failed to produce an orthomosaic. 
We observed this failure mode for numerous reconstructions with low-light conditions, suggesting that mean fusion is a more resilient method compared to other approaches. 

Qualitative results can be seen
in Fig.~\ref{fig:qualitative}. We observe that areas with artificial
illumination reconstruct well in the RGB images, but the rest of the image is
quite dark. On the other hand, the mean fusion method is able to increase the
overall brightness of the image, improving the level of detail in parking lots
and sidewalks. However, the fusion method increases the overall level of noise
in the image, producing artifacts in the reconstruction.

It was, however, surprising to us that the RGB camera performed well in these very low light situations. This behavior can be explained thanks to the high quantum efficiency of the sensor used (60\% at $525\text{nm}$), and the maximum exposure time of $15\,\text{ms}$.

\fer{In our results, we observe that frames synthesized from events are able to overcome the limitations of RGB images given their high dynamic range. Nevertheless, pure event representations struggle with high texture complexities such as grassy areas. We believe this is due to the limitations of the sensor, which generates a significant number of events that may saturate its bandwidth.}
\section{Conclusion}
In this work, we presented a method for high-altitude orthomapping using event cameras.
To our knowledge, this is the first work that leverages the characteristics of
event cameras to overcome the issues of CMOS-based sensors in this setting.
We demonstrated that our method enables the creation of orthomosaics in low-light conditions, as well as qualitative improvements in high dynamic range areas. 

Given the success of this work, there are exciting possibilities for future work.
For example, the experiment area for this paper offered a limited amount of high dynamic range scenarios, and thus the overall \gls{psnr} and \gls{ssim} metrics performed better than event-based reconstructions.
We expect event based orthomosaic mapping to perform better in more challenging scenarios, such as forests at noon.  
Furthermore, in this work, we leveraged traditional image processing pipelines to produce the orthomosaics, which relies on FLANN features for matching~\cite{muja2009flann}. 
However, feature tracking in the event space may provide more resilient
tracking, particularly in low-light conditions~\cite{messikommer2023data,
gehrig2018asynchronous}.  
Additionally, although we relied on \gls{gnss} priors to generate the reconstruction, there are promising future directions using event-based SLAM approaches for orthomapping reconstruction, such as the ones
proposed by~\cite{shuang2024cmax}.
Finally, we used traditional methods for pansharpening. Consequently, we observed that the noise in the fused images was higher than the base RGB. 
Generative methods have outperformed traditional methods in satellite imaging and one possibility is training these generative methods on event data to yield sharper reconstructions.

\longshot{Additionally, the
use of an event camera allows motion segmentation in the image plane, which can
be used to track independently moving objects.}

%\section*{APPENDIX}

\section*{ACKNOWLEDGMENTS}
The authors would like to thank Alex Zhou and Jeremy Wang for their help maintaining the  platforms and manufacturing the hardware components of the \gls{uav}, and Benedict Onyekwe for his help routing the synchronization board. We would like to acknowledge the invaluable work of Bernd Pfrommer for maintaining the open-source ROS drivers used for data collection and post-processing for both event and RGB cameras, as well as his valuable suggestions to generate event reconstructions. We finally would like to thank Victoria Edwards and Mariana Quesada for reviewing this manuscript.

\addtolength{\textheight}{-8.7cm}   % This command serves to balance the column lengths on the last page of the document manually.
\bibliographystyle{IEEEtran}
\bibliography{IEEEabrv, literature}

\begin{thebibliography}{10}
\providecommand{\url}[1]{#1}
\csname url@samestyle\endcsname
\providecommand{\newblock}{\relax}
\providecommand{\bibinfo}[2]{#2}
\providecommand{\BIBentrySTDinterwordspacing}{\spaceskip=0pt\relax}
\providecommand{\BIBentryALTinterwordstretchfactor}{4}
\providecommand{\BIBentryALTinterwordspacing}{\spaceskip=\fontdimen2\font plus
\BIBentryALTinterwordstretchfactor\fontdimen3\font minus
  \fontdimen4\font\relax}
\providecommand{\BIBforeignlanguage}[2]{{%
\expandafter\ifx\csname l@#1\endcsname\relax
\typeout{** WARNING: IEEEtran.bst: No hyphenation pattern has been}%
\typeout{** loaded for the language `#1'. Using the pattern for}%
\typeout{** the default language instead.}%
\else
\language=\csname l@#1\endcsname
\fi
#2}}
\providecommand{\BIBdecl}{\relax}
\BIBdecl

\bibitem{valenzuela2019basic}
A.~Q. Valenzuela and J.~C.~G. Reyes, ``{Basic spatial resolution metrics for
  satellite imagers},'' \emph{IEEE Sensors Journal}, vol.~19, no.~13, pp.
  4914--4922, 2019.

\bibitem{jacobsen2005high}
K.~Jacobsen \emph{et~al.}, ``{High resolution satellite imaging systems-an
  overview},'' \emph{Photogrammetrie Fernerkundung Geoinformation}, vol. 2005,
  no.~6, p. 487, 2005.

\bibitem{pix4dProfessionalPhotogrammetry}
``{{P}rofessional photogrammetry and drone mapping software --- pix4d.com},''
  \url{https://www.pix4d.com/}, [Accessed 01-09-2024].

\bibitem{opendronemapDroneMapping}
``{{D}rone {M}apping {S}oftware - {O}pen{D}rone{M}ap™ ---
  opendronemap.org},'' \url{https://www.opendronemap.org/}, [Accessed
  01-09-2024].

\bibitem{opendronemapTutorialsx2014}
``{{T}utorials; {O}pen{D}rone{M}ap 3.5.3 documentation ---
  docs.opendronemap.org},'' \url{https://docs.opendronemap.org/tutorials/},
  [Accessed 01-09-2024].

\bibitem{li2024ers}
X.~Li, S.~Cheng, Z.~Zeng, C.~Zhao, and C.~Fan, ``{ERS-HDRI: Event-Based Remote
  Sensing HDR Imaging},'' \emph{Remote Sensing}, vol.~16, no.~3, p. 437, 2024.

\bibitem{escudero2023enabling}
N.~Escudero, M.~W. Hardt, and G.~Inalhan, ``{Enabling UAVs night-time
  navigation through Mutual Information-based matching of event-generated
  images},'' in \emph{2023 IEEE/AIAA 42nd Digital Avionics Systems Conference
  (DASC)}.\hskip 1em plus 0.5em minus 0.4em\relax IEEE, 2023, pp. 1--10.

\bibitem{scheerlinck2018continuous}
C.~Scheerlinck, N.~Barnes, and R.~Mahony, ``{Continuous-time intensity
  estimation using event cameras},'' in \emph{Asian Conference on Computer
  Vision}.\hskip 1em plus 0.5em minus 0.4em\relax Springer, 2018, pp. 308--324.

\bibitem{haoyu2020learning}
C.~Haoyu, T.~Minggui, S.~Boxin, W.~YIzhou, and H.~Tiejun, ``{Learning to deblur
  and generate high frame rate video with an event camera},'' \emph{arXiv
  preprint arXiv:2003.00847}, 2020.

\bibitem{messikommer2022multi}
N.~Messikommer, S.~Georgoulis, D.~Gehrig, S.~Tulyakov, J.~Erbach,
  A.~Bochicchio, Y.~Li, and D.~Scaramuzza, ``{Multi-bracket high dynamic range
  imaging with event cameras},'' in \emph{Proceedings of the IEEE/CVF
  conference on computer vision and pattern recognition}, 2022, pp. 547--557.

\bibitem{gallego2020event}
G.~Gallego, T.~Delbr{\"u}ck, G.~Orchard, C.~Bartolozzi, B.~Taba, A.~Censi,
  S.~Leutenegger, A.~J. Davison, J.~Conradt, K.~Daniilidis \emph{et~al.},
  ``{Event-based vision: A survey},'' \emph{IEEE transactions on pattern
  analysis and machine intelligence}, vol.~44, no.~1, pp. 154--180, 2020.

\bibitem{rebecq2019high}
H.~Rebecq, R.~Ranftl, V.~Koltun, and D.~Scaramuzza, ``{High speed and high
  dynamic range video with an event camera},'' \emph{IEEE transactions on
  pattern analysis and machine intelligence}, vol.~43, no.~6, pp. 1964--1980,
  2019.

\bibitem{brandli2014real}
C.~Brandli, L.~Muller, and T.~Delbruck, ``{Real-time, high-speed video
  decompression using a frame-and event-based DAVIS sensor},'' in \emph{2014
  IEEE International Symposium on Circuits and Systems (ISCAS)}.\hskip 1em plus
  0.5em minus 0.4em\relax IEEE, 2014, pp. 686--689.

\bibitem{pfrommer2024recon}
B.~Pfrommer, ``{simple\_image\_recon - simple image reconstruction for an event
  based camera},'' \url{https://github.com/berndpfrommer/simple_image_recon},
  [Accessed 06-09-2024].

\bibitem{bisulco2021fast}
A.~Bisulco, F.~Cladera, V.~Isler, and D.~D. Lee, ``{Fast Motion Understanding
  with Spatiotemporal Neural Networks and Dynamic Vision Sensors},'' in
  \emph{2021 IEEE International Conference on Robotics and Automation (ICRA)},
  2021, pp. 14\,098--14\,104.

\bibitem{Rebecq19cvpr}
H.~Rebecq, R.~Ranftl, V.~Koltun, and D.~Scaramuzza, ``{Events-to-Video:
  Bringing Modern Computer Vision to Event Cameras},'' \emph{{IEEE} Conf.
  Comput. Vis. Pattern Recog. (CVPR)}, 2019.

\bibitem{Chaney_2023_CVPR}
K.~Chaney, F.~Cladera, Z.~Wang, A.~Bisulco, M.~A. Hsieh, C.~Korpela, V.~Kumar,
  C.~J. Taylor, and K.~Daniilidis, ``{M3ED: Multi-Robot, Multi-Sensor,
  Multi-Environment Event Dataset},'' in \emph{Proceedings of the IEEE/CVF
  Conference on Computer Vision and Pattern Recognition (CVPR) Workshops}, June
  2023, pp. 4015--4022.

\bibitem{osadcuks2020clock}
V.~Osadcuks, M.~Pudzs, A.~Zujevs, A.~Pecka, and A.~Ardavs, ``{Clock-based time
  synchronization for an event-based camera dataset acquisition platform},'' in
  \emph{2020 IEEE International Conference on Robotics and Automation
  (ICRA)}.\hskip 1em plus 0.5em minus 0.4em\relax IEEE, 2020, pp. 4695--4701.

\bibitem{liu2022large}
X.~Liu, G.~V. Nardari, F.~Cladera, Y.~Tao, A.~Zhou, T.~Donnelly, C.~Qu, S.~W.
  Chen, R.~A.~F. Romero, C.~J. Taylor, and V.~Kumar, ``{Large-Scale Autonomous
  Flight With Real-Time Semantic SLAM Under Dense Forest Canopy},'' \emph{IEEE
  Robotics and Automation Letters}, vol.~7, no.~2, pp. 5512--5519, 2022.

\bibitem{flirCamDriver}
B.~Pfrommer, ``{flir\_camera\_driver - ROS Teledyne FLIR camera drivers},''
  \url{https://github.com/ros-drivers/flir_camera_driver/tree/humble-devel},
  [Accessed 16-09-2024].

\bibitem{hurliman2024mcap}
J.~Hurliman, ``{MCAP: A Next-Generation File Format for ROS Recording},''
  \url{http://download.ros.org/downloads/roscon/2022/MCAP%20A%20Next-Generation%20File%20Format%20for%20ROS%20Recording.pdf},
  2022, [Accessed 02-09-2024].

\bibitem{folk2011overview}
M.~Folk, G.~Heber, Q.~Koziol, E.~Pourmal, and D.~Robinson, ``{An overview of
  the HDF5 technology suite and its applications},'' in \emph{Proceedings of
  the EDBT/ICDT 2011 workshop on array databases}, 2011, pp. 36--47.

\bibitem{rehder2016extending}
J.~Rehder, J.~Nikolic, T.~Schneider, T.~Hinzmann, and R.~Siegwart, ``{Extending
  kalibr: Calibrating the extrinsics of multiple IMUs and of individual
  axes},'' in \emph{2016 IEEE International Conference on Robotics and
  Automation (ICRA)}.\hskip 1em plus 0.5em minus 0.4em\relax IEEE, 2016, pp.
  4304--4311.

\bibitem{vivione2015critical}
G.~Vivone, L.~Alparone, J.~Chanussot, M.~Dalla~Mura, A.~Garzelli, G.~A.
  Licciardi, R.~Restaino, and L.~Wald, ``{A Critical Comparison Among
  Pansharpening Algorithms},'' \emph{IEEE Transactions on Geoscience and Remote
  Sensing}, vol.~53, no.~5, pp. 2565--2586, 2015.

\bibitem{muja2009flann}
M.~Muja and D.~Lowe, ``{Flann-fast library for approximate nearest neighbors
  user manual},'' \emph{Computer Science Department, University of British
  Columbia, Vancouver, BC, Canada}, vol.~5, no.~6, 2009.

\bibitem{messikommer2023data}
N.~Messikommer, C.~Fang, M.~Gehrig, and D.~Scaramuzza, ``{Data-driven feature
  tracking for event cameras},'' in \emph{Proceedings of the IEEE/CVF
  Conference on Computer Vision and Pattern Recognition}, 2023, pp. 5642--5651.

\bibitem{gehrig2018asynchronous}
D.~Gehrig, H.~Rebecq, G.~Gallego, and D.~Scaramuzza, ``{Asynchronous,
  photometric feature tracking using events and frames},'' in \emph{Proceedings
  of the European Conference on Computer Vision (ECCV)}, 2018, pp. 750--765.

\bibitem{shuang2024cmax}
S.~Guo and G.~Gallego, ``{CMax-SLAM: Event-Based Rotational-Motion Bundle
  Adjustment and SLAM System Using Contrast Maximization},'' \emph{IEEE
  Transactions on Robotics}, vol.~40, pp. 2442--2461, 2024.

\end{thebibliography}

\end{document}